\documentclass[runningheads]{llncs}
\usepackage{graphicx}
\usepackage{booktabs}
\usepackage[utf8]{inputenc}
\usepackage[T1]{fontenc}
\usepackage{bm}
\usepackage{amsmath,amsfonts}
\usepackage{lineno}%
\usepackage{makecell}

\newcommand{\bi}{\begin{itemize}}
\newcommand{\ei}{\end{itemize}}
\newcommand{\ba}{\begin{array}}
\newcommand{\ea}{\end{array}}

\usepackage{xspace}

\newcommand{\bmx}[0]{\begin{bmatrix}}
\newcommand{\emx}[0]{\end{bmatrix}}

\newif\ifarxiv
\arxivfalse 

\begin{document}
\title{Action valuation of on- and off-ball soccer players \\ based on multi-agent deep reinforcement learning}
%
\titlerunning{Valuation of multiple soccer players via deep reinforcement learning}
%

\ifarxiv
\else
\author{Hiroshi Nakahara\inst{1} \and Kazushi Tsutsui\inst{1} \and Kazuya Takeda\inst{1} \and Keisuke Fujii\inst{1,2,3}}
%
\institute{Graduate School of Informatics, Nagoya University, Nagoya, Japan. \and
Center for Advanced Intelligence Project, RIKEN, Suita, Japan.  \and 
PRESTO, Japan Science and Technology Agency, Saitama, Japan. 
}
\authorrunning{H. Nakahara et al.}
\vspace{-11pt}

\fi
\maketitle              
\begin{abstract}
\vspace{-11pt}
Analysis of invasive sports such as soccer is challenging because the game situation changes continuously in time and space, and multiple agents individually recognize the game situation and make decisions. Previous studies using deep reinforcement learning have often considered teams as a single agent and valued the teams and players who hold the ball in each discrete event. Then it was challenging to value the actions of multiple players, including players far from the ball, in a spatiotemporally continuous state space. In this paper, we propose a method of valuing possible actions for on- and off-ball soccer players in a single holistic framework based on multi-agent deep reinforcement learning. We consider a discrete action space in a continuous state space that mimics that of Google research football and leverages supervised learning for actions in reinforcement learning. 
In the experiment, we analyzed the relationships with conventional indicators, season goals, and game ratings by experts, and showed the effectiveness of the proposed method.
Our approach can assess how multiple players move continuously throughout the game, which is difficult to be discretized or labeled but vital for teamwork, scouting, and fan engagement.

\vspace{-6pt}
\end{abstract}
\vspace{-25pt}
\section{Introduction}
\vspace{-9pt}
\label{sec:introduction}
With advanced data analysis in sports, tactical planning, player evaluation, and coaching methods based on data have recently become available.
The data analytics of dynamic movements in team sports such as soccer is considered challenging because game situations are continuous in time and space, and multiple agents individually recognize the game situations and make decisions.
In particular, each action's value has been quantified based on the strength of its association with scores for on-ball players (i.e., with a ball).
In previous work using supervised learning, machine learning models were used to compute an action's value by predicting whether scores or other events occur or not in the following actions \cite{Decroos19,Toda2021,umemoto2022location,yeung2023transformer}.
In these frameworks, it would be difficult to consider possible (i.e., counterfactual) actions as time goes back from a goal or other events. 
To value on-ball actions in terms of obtaining rewards (e.g., goals), there have been studies using reinforcement learning (RL) \cite{Liu2018,Liu2020,Schulte2015,rahimian2022beyond}.
These works typically consider teams as a single agent and valuate an on-ball player or a team in irregularly occurring events (e.g., passes and shots).
Considering the nature of soccer, it should be modeled as a continuous space, and valuate actions without event labels including on- and off-ball plays.

Although most studies focus on the evaluation of on-ball players, players can indirectly contribute to scoring even when they are off-ball (i.e., without a ball) which is a large part of their playing time (e.g., approximately 87 min of 90 min \cite{Fernandez18}). 
A previous work \cite{Spearman18} estimated the value of the state from the positional information of the ball and players based on a rule-based model called Off-Ball Scoring Opportunities (OBSO).
However, valuing other attacking players who do not receive the ball and reflecting the valuation of several possible actions in the state value is challenging. 
Another study \cite{teranishi2022evaluation} quantified every off-ball player's impact on scores in terms of the difference between predicted and real player movements.
The method quantitatively values only a single player's contribution once through their predicted movement trajectories.
Thus it is challenging to calculate the contributions of multiple players at each time stamp using a comprehensive learning-based framework.

In this paper, we propose a valuation method of on- and off-ball soccer players in a single holistic framework based on multi-agent deep RL. 
Specifically, we consider a discrete action space in a continuous state space that mimics that of the Google research football (GFootball) \cite{kurach2020google} and leverage supervised learning for actions in RL. 
Based on a deep RL model with a discrete action space, we estimate the possible action value of multiple players in real games, including those far from the ball.
The proposed network estimates state-action values (i.e., Q-values) based on the game states (e.g., player and ball locations) and actions (e.g., shot and pass). 
For the off-ball action, we define the directions of movement as actions. 
For player valuations, our method can compute the overall contribution of each player during the attack by aggregating the Q-values in the RL model.
The proposed method enables us to comprehensively value on- and off-ball actions, which makes it possible to compare the contributions of the two types of actions, and thus provides important information for understanding the characteristics of players. 
Assessing the movements of all players for teammates is important for building teamwork, assessment of players' salaries, recruitment, and scouting.

In summary, our main contributions were as follows.
(1) We propose a valuation method of on- and off-ball soccer players in a single holistic framework based on multi-agent deep RL. 
(2) We consider a discrete action space in a continuous state space that mimics GFootball and leverages supervised learning for actions in RL. Furthermore, since the proposed method can counterfactually compute possible action values that were not chosen in the actual game, it can be also used for valuating counterfactual actions. 
(3) In the experiment, we analyzed the relationships with conventional indicators, season goals, and game ratings by experts, and showed the effectiveness of the proposed method.
Our approach can assess how multiple players move continuously throughout the game, which is difficult to be discretized or labeled but vital for teamwork, scouting, and fan engagement.
The remainder of this paper is structured as follows.
First, we describe our methods in Section \ref{sec:method} and present experimental results in Section \ref{sec:experiment}.
Next, we overview the related works in Section \ref{sec:related} and conclude this paper in Section \ref{sec:conclusion}.

\vspace{-10pt}
\section{Method}
\label{sec:method}
\vspace{-5pt}
In this section, we describe our problem setting and propose our RL framework.
Then we describe the dataset and preprocessing, and valuation framework.

\vspace{-7pt}
\subsection{Problem setting}
\vspace{-3pt}
In this study, we aim to value players by computing state-action values (i.e., Q-values) using an RL framework. 
For simplicity, here we consider independent multi-agent RL and then omit the agent index.

The RL model in this study consists of three components: state $s$, action $a$, and reward $r$. 
At the time $t$, when an action $a_t$ is chosen in a state $s_t$, a reward $r_{t+1}$ is given.
In on-policy RL, the agent learns a policy $\pi$ that can maximize $\sum_{t=1}^{T} \gamma^{t} r_{t}$, where $\gamma \in [0,1]$ is the discount factor and $T$ is the time horizon (hereafter, we consider the case of $\gamma = 1$ \cite{Liu2020} with no discount for simplicity).

For the state $s_t$, we used 2-dimensional position coordinates and velocities data of the players ($22$) and the ball ($23 \times 2 \times 2$ dimensions). 
Inspired by 19 actions defined in GFootball \cite{kurach2020google}, we selected 14 actions of the attacking players: movement actions defined as $8$ different movement directions ($8$ directions in $45$ degree increments), idle, starting and stopping a sprint, the release of the movement direction, passing, and shooting.
The passing and shooting actions were defined by the labels (event data) in the dataset described later. 
Movement directions were computed based on the player's velocity direction. 
Other, idle, starting and stopping a sprint, and release of the movement direction were computed based on the stopping (0.1 m/s) and sprint thresholds (24 km/h). 
Regarding on-ball players' actions, the dataset included other behavior labels (e.g., dribbling and trapping), which were not used in this study for simplicity. 

For the reward $r$, we added the following three values at the last time $T$ of the sequence of attacks.  
(1) Goal: $1$ if a series of attacks end in a score, and $0$ otherwise, 
(2) Expected score with no goal: the EPV (expected possession value) \cite{Fernandez2019}, which is an indicator of the probability of scoring based on the ball position coordinates ($x, y$) at time $T$,
(3) Conceding: $-1$ if a goal is scored by an opponent's attack immediately after a sequence of attacks, and $0$ otherwise. 
For the rewards at $t$ other than time $T$ ($t < T$), we used $0$ in this study. 

\begin{figure}[t]
  \centering
  \includegraphics[width=12cm]{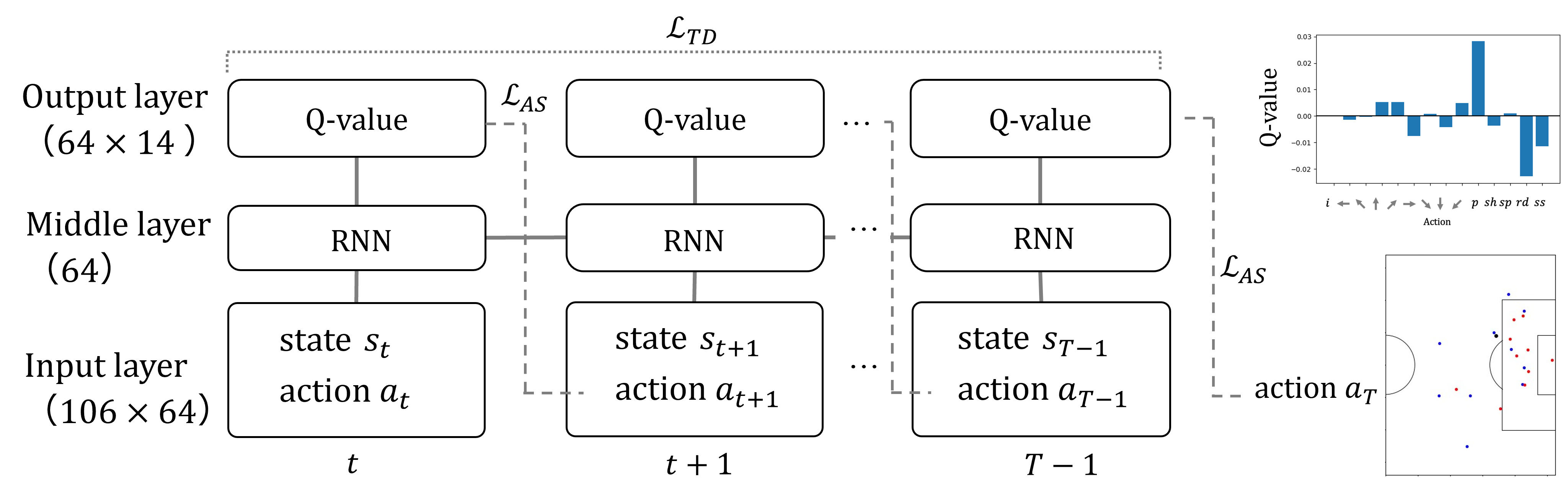}
  \caption{Our deep RL model with action supervision.}
  \label{fig:architecture}
  \vspace{-10pt}
\end{figure}

\vspace{-7pt}
\subsection{Deep reinforcement learning with action supervision}
\vspace{-3pt}
Next, we explain our RL algorithm with action supervision.  
We compute state-action values (Q-value) for the valuation of plays, which has been used in previous sports studies \cite{Liu2018,Liu2020,ding2022deep,rahimian2022beyond}.
In this framework, the optimal state-action value function $Q^*(s_t,a_t)$ is determined by solving the Bellman equation as follows:
\begin{eqnarray}
Q^*(s_t, a_t) &=& E_{s_{t+1}, a_{t+1}} [r_{t+1} + Q(s_{t+1}, a_{t+1})|s_t, a_t] \\
&=& \sum_{r_{t+1}} P(r_{t+1}|s_t, a_t)r_{t+1} + \sum_{\substack{s_{t+1} \\ a_{t+1}}} P(s_{t+1}, a_{t+1}|s_t, a_t)Q(s_{t+1}, a_{t+1}),\label{eq:q-function}
\end{eqnarray}
where $Q(s_t, a_t)$ refers to the Q-value of taking action $a$ in state $s$. 
In other words, the Q-value of taking action $a_t$ in state $s_t$ is defined as the sum of the expected reward when the sequence of attacks ends at time $t$ and the expected Q-value at time $t+1$. 

To train the RL model, following by \cite{Liu2018,Liu2020}, we adopt a temporal difference (TD) prediction method called SARSA (state-action-reward-state-action), which is a model-free and on-policy RL algorithm, with recurrent neural network (RNN). 
As the input sequences, we used a matrix of $T$ frames of state (92 dim.) and the one-hot action for each agent (14 dim.) as input.   
A conceptual diagram of the architecture is shown in Fig.  \ref{fig:architecture}.
We used GRUs (gated recurrent units) as RNNs. 
The middle layer was a single layer with $64$-dimensional neurons and the activation function was ReLU.
In the output layer, we output the Q-values for each player for each action (14 dim.). 
To estimate Q values based on Eq. (\ref{eq:q-function}), we compute the following TD loss:
\begin{equation}
\mathcal{L}_{TD} = \sum_{t \in T} (r_{t+1} + Q(s_{t+1}, a_{t+1}) - Q(s_t, a_t))^2.
\end{equation}

In addition, we introduce a supervised loss for actions because the above approach might lead to insufficient learning of all possible actions and it would be necessary to restrict the state spaces from the limited data. 
Based on the discussion in \cite{hester2018deep,fujii2023adaptive}, we propose a simple action supervision loss represented by the cross-entropy of softmax values of the Q-function such that

\begin{equation}
\mathcal{L}_{AS} = -\sum_{t \in T} \mathbf{a}_{t}\cdot 
\log
\left(\mathrm{softmax}(\mathbf{Q}_{s_t})
\right), 
\label{eq:simplesupervised}
\end{equation}
where $\mathbf{a}_t \in \{0, 1\}^{|A|}$ (i.e., one-hot vector of actions), $|A|$ is the size of action space, $\mathbf{Q}_{s_t} = [ Q(s_t,a_t=1), ..., Q(s_t,a=|A|) ]$, and the $\log$ applies element-wise. 
This loss aims to maximize the Q-function values for the action of the data.
Note that we assume that the action of the actual players is better than random selection.
This is an inductive bias for efficient learning to estimate Q-values for all possible actions but the main loss is $\mathcal{L}_{TD}$ and the weight of $\mathcal{L}_{AS}$ should be much smaller than $\mathcal{L}_{TD}$.

We also add an $L_1$ regularization loss applied to the weights and biases of the network to help prevent over-fitting on the relatively small demonstration dataset.
The total loss in our training is a
combination of three losses:
\begin{equation}
  \mathcal{L}_{total} = \mathcal{L}_{TD} + {\lambda_1}\mathcal{L}_{AS}+ {\lambda_2}\mathcal{L}_{L_1}.
  \label{eq:losses}
\end{equation}
We set $\lambda_1 = 0.01$ and $\lambda_2 = 0.1$, which control the weighting among the losses.
We trained the models using the Adam optimizer \cite{Kingma15} with default parameters.

\vspace{-5pt}
\subsection{Dataset and preprocessing}
In this study, we used 54 games data in the Meiji J1 League (a professional soccer league in Japan) 2019 season including all 34 games data of Yokohama F Marinos to perform specific player-level evaluations in limited data.  
Note that currently the tracking data for all players and timesteps were not publicly shared in such amounts. 
The dataset includes event data (i.e., labels of actions, e.g., passing and shooting, recorded at 30 Hz and the simultaneous xy coordinates of the ball) and tracking data (i.e., xy coordinates of all players recorded at 25 Hz) provided by Data Stadium Inc.
The season goals for each player in each match were collected from \cite{Score}.
The ratings by experts in each match \cite{Rating} were also used, which was scored in 0.5-point increments with a maximum of 10 points.

As a preprocessing step, we first split a sequence of a game (usually about $90$ minutes) into possessions (i.e., sequences of attacks) from the beginning of the possession (ball recovery) to the end of the possession (ball loss or goal) as an input of RL models. 
The minimum number of frames was $50$, and the maximum number of frames was $T=300$ (equivalent to $30$ seconds). 
The tracking data were down-sampled to 10 Hz based on \cite{fujii2020policy,teranishi2022evaluation}.
We analyzed 10 attacking players (without a goalkeeper), i.e., constructed 10 RL agent models.

\subsection{Validation of the model and valuation of players}
In the experiment, we first validated the RL model and then analyzed the valuations of professional soccer players.
To train the model, we used attack sequences other than those of Yokohama F. Marinos ($1669$ and $186$ sequences) as the training and validation data.
To value the players, we used attack sequences of Yokohama F. Marinos ($1236$ series) as the test data. 
For simplicity, only attacks within the attacking third were used.
We used attack sequences other than those of Yokohama F. Marinos ($1669$ series) as the training data, and attack sequences of Yokohama F. Marinos ($1236$ series) as the test data for valuating the actions.

\begin{figure}[t]
  \centering
  \includegraphics[width=12cm]{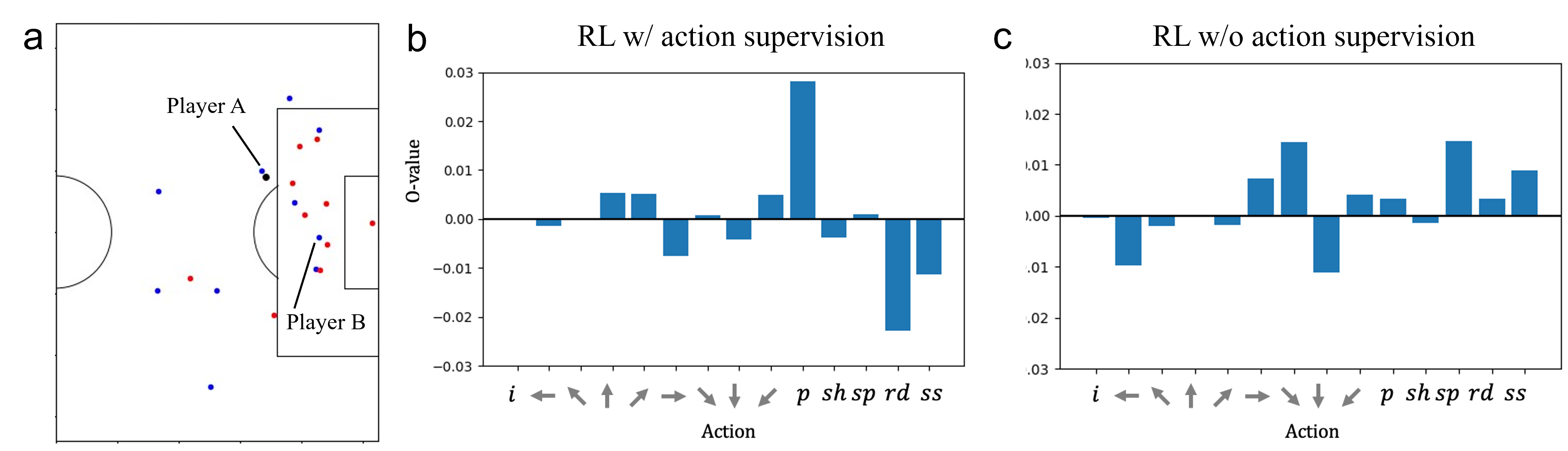}
  \caption{Example of estimated Q-values. (a) shows an example attack. Blue, red, and black indicate the attacker, defender, and the ball, respectively (note that the ball is over a defender). (b) and (c) show the Q-value of player A for each action using the RL model with and without action supervision, respectively. Note that $i, p, sh, sp, rd, ss$ correspond to idle, pass, shot, sprint, decelerate, and sprint end, respectively, and arrows correspond to the direction of movement.
  }
  \label{fig:q_output_example}
  \vspace{-10pt}
\end{figure}

\section{Results}
\label{sec:experiment}
In this section, we validated the RL model, and then analyzed the valuations of professional soccer players.
\vspace{-5pt}
\subsection{Validation of our RL model} 
Here we compared our RL models with and without action supervision (i.e., $\lambda_1 = 0$ and other conditions were the same).  
We quantitatively compared the TD loss in RL $\mathcal{L}_{TD}$ and action supervision loss $\mathcal{L}_{AS}$.
The $\mathcal{L}_{TD}$ with supervision ($0.0034 \pm 0.0001$; the mean and standard deviation of test samples) was smaller than that without supervision ($0.0063 \pm 0.0004$), but $\mathcal{L}_{AS}$ with supervision ($3.9550 \pm 0.0005$) was slightly larger than that without supervision ($3.9407 \pm 0.0005$).
Considering the fact that the reward scale is $[-1, 1]$, the learning of Q-values was considered to be within the acceptable range for both models.
It should be noted that we can quantitatively compare the optimization results (i.e., losses) but can only qualitatively investigate the effectiveness of the model in terms of modeling of soccer agents. 
In particular, the action supervision may require careful interpretation, because much less supervision (e.g., $\lambda_1 \ll 0.01$) would lead to insufficient learning of counterfactual action values, whereas much more supervision (e.g., $\lambda_1 \gg 0.01$) may overfit to the actual actions and would not consider counterfactual actions.
Then we carried out a qualitative analysis as described below. 

Next, an example of the Q-value output is shown. 
Fig. \ref{fig:q_output_example}a shows the coordinates of the player and the ball in the frame to be valuated.
In this case, player A in the actual game passed the ball to player B, who was considered to have more space for a shot.
Figs. \ref{fig:q_output_example}b and c show the Q-value of player A for each action using the RL model with and without action supervision, respectively. 
As shown in Fig. \ref{fig:q_output_example}b, our model with action supervision indicates that the Q-value of a pass was higher than those of others, suggesting that passing may produce a more favorable result rather than other actions (e.g., a shot) in this case.
In contrast, in Fig. \ref{fig:q_output_example}c, our model without action supervision shows more distributed Q-values closer to zero and small value of the pass action.
Quantitatively, the average on-ball Q-values with supervision ($0.0137 \pm 2.597 \times 10^{-4}$) were larger than those without supervision ($0.0008 \pm 4.431 \times 10^{-5}$), and the average off-ball Q-values with supervision ($0.0008 \pm 0.0035$) were also larger than those without supervision ($-7.123 \pm 0.0018$). 
In particular, our approach with supervision emphasized the on-ball valuation, which was similar to the conventional valuation of the players, but in some cases, a fairer valuation including off-ball situations may be required. 
Although which models were best in a practical sense cannot be determined from the data, for validation of the model output, we mainly show the results of the model with action supervision based on the above verification.

\begin{figure}[t]
  \centering
  \includegraphics[width=12cm]{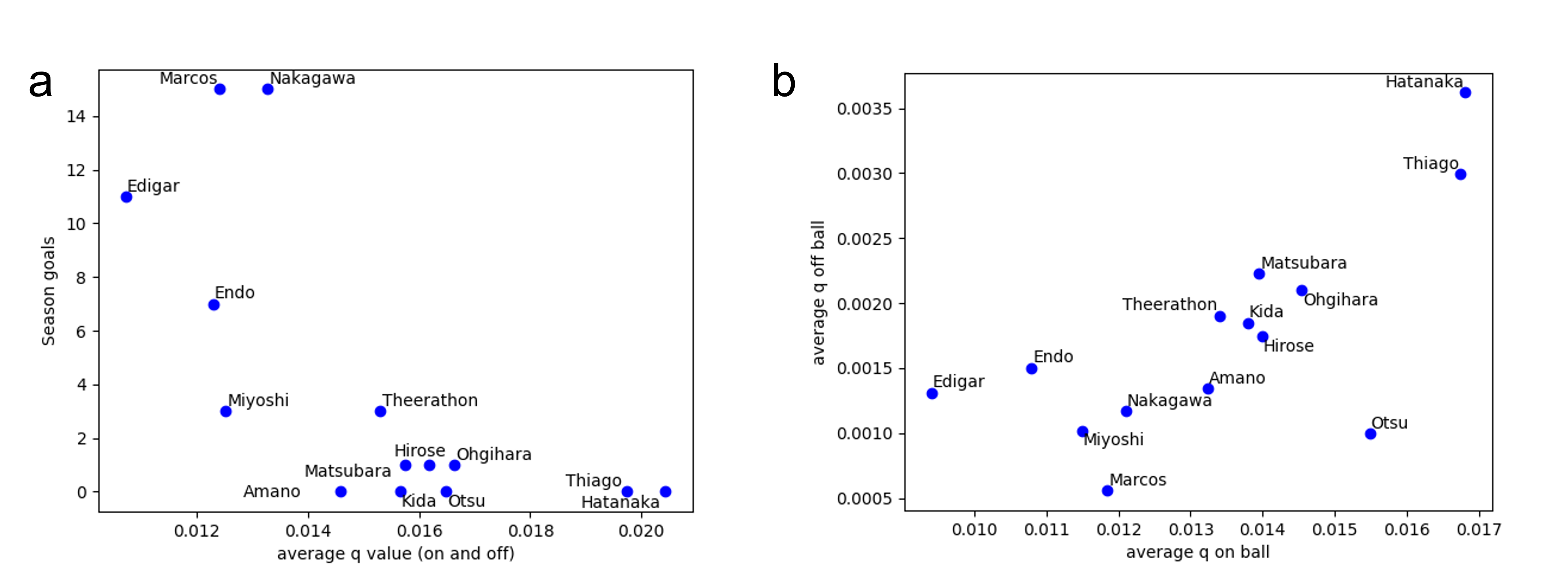}
  \caption{The relationship between (a) average Q-values computed by the proposed method and season goals and (b) average Q-values of on- and off-ball states.}
  \label{fig:goal-on-off-Q}
  \vspace{-10pt}
\end{figure}

\vspace{-5pt}
\subsection{Comparison with conventional indicators and player valuation}
\vspace{-5pt}
Since there is no true value for the Q-value, the performance of the model cannot be directly validated. 
Here we show the usefulness of the proposed indicators by investigating the relationship with the existing indicators. 
Specifically, we explain the relationships between the average Q-values computed by the proposed method and four indicators: season goals, ratings by experts, OBSO \cite{Spearman18}, and creating off-ball scoring opportunity (C-OBSO) \cite{teranishi2022evaluation}. 
OBSO was used to estimate the value of the state based on the rule-based model, which basically values attacking players who will receive the ball. 
C-OBSO quantifies the off-ball player's impact on scores created by the difference between predicted and real player movements.
Spearman's rank correlation coefficient (hereafter denoted as $\rho$) was used as the correlation coefficient, and the players to be valuated were limited to those who played at least $10$ games. 
Since the sample size was small ($N=14$) in the correlation analysis, the $\rho$ value was used as an effect size for evaluation (criteria are based on \cite{guilford1950fundamental}), rather than the $p$-value.

First, the relationship between the season goals and the average Q-values obtained by the proposed method is shown in Fig. \ref{fig:goal-on-off-Q}a. 
There was a high negative correlation between them ($\rho = -0.761$) whereas many players had 0 or 1 total goal. 
In the season, four players (Nakagawa, Marcos, Thiago, and Kida) were selected as the best 11 players award in the league. 
Among these four players, the proposed method showed higher values for a defender (Thiago), midfielder (Kida), and forward players (Nakagawa and Marcos) in this order.
Considering the high negative correlation, our indicator tended to value defensive players who provided many passes, because Thiago ranked 6th in terms of the players with the most passes in the league in spite of being a defender.
Similarly, Hatanaka, who was a defender ranked 1st in our indicator, also ranked 2nd in terms of most passes in the league.
The results suggest that their ability indirectly contributing to the goal with many passes and other off-ball movements can be reflected in our indicator, rather than goals.

Next, we compared the contributions of on- and off-ball plays. 
In Fig. \ref{fig:goal-on-off-Q}, there was a moderate positive correlation between average on- and off-ball Q-values ($\rho = 0.618$), suggesting that our indicator may tend to value similar play style between on- and off-ball situations. 
Because of a similar tendency, in the following analysis, we also used the Q-values including on- and off-ball cases. 

\begin{figure}[t]
  \centering
  \includegraphics[width=14cm]{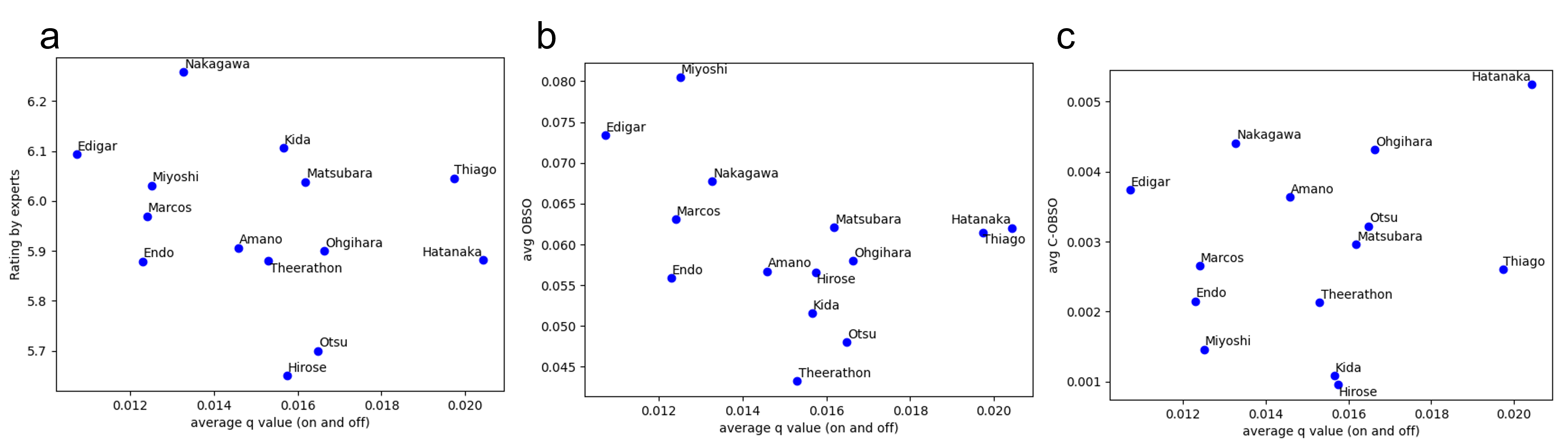}
  \caption{The relationship between average Q-values by the proposed method and (a) the average rating by the experts, (b) OBSO \cite{Spearman18}, and (c) C-OBSO \cite{teranishi2022evaluation}.}
  \vspace{-10pt}
  \label{fig:metrics_Q}
\end{figure}

Next, the relation between the average rating by the experts and the average Q-value is shown in Fig. \ref{fig:metrics_Q}a. 
There was a low negative correlation between the two indicators ($\rho = -0.218$). 
This result may be related to the tendency that our indicator tended to value defense players who provided many passes, and there was a significant correlation between the goals and this subjective rating in a previous work \cite{teranishi2022evaluation}.

Finally, we compared our results with existing off-ball indicators such as OBSO \cite{Spearman18} and C-OBSO \cite{teranishi2022evaluation}.
Figure \ref{fig:metrics_Q}b shows a low negative correlation with average OBSO ($\rho = -0.305$). 
Since OBSO values attacking players who will receive the ball, the tendency was similar to the correlation with the season goal. 
OBSO can also value the players' all time stamps, thus the offensive players can be valued separately based on both OBSO and our Q-values. 
For example, regarding Miyoshi (a midfielder) and Theerathon (a defender), who had three season goals, Miyoshi's performance can be highly valued in terms of off-ball movement to receive the ball (i.e., OBSO), and Theerathon's can be highly valued in terms of passing and other off-ball movements (i.e., our method). 

Figure \ref{fig:metrics_Q}c shows no correlation with average C-OBSO ($\rho = 0.182, p > 0.05$). 
Compared with C-OBSO \cite{teranishi2022evaluation}, which can value forward players such as Edigar and Nakagawa more than our Q-values, our Q-values tended to value midfielders and defenders such as Kida (0 season goals) and Hirose (1 season goal) indirectly contributing to the goals. 
From these results, we clarified the properties of our Q-values compared with those of conventional indicators.

\section{Related work}
\label{sec:related}
\vspace{-5pt}
In team sports tactical behaviors, agents’ behavioral process bears resemblance to the RL framework (e.g., \cite{fujii2021data}).
Due to various challenges in modeling the entire framework from data, two approaches can be adopted: estimating related variables and functions from data as a sub-problem (i.e., the inverse approach) and constructing a model (e.g., an RL model) to generate data in a virtual environment (i.e., the forward approach such as \cite{kurach2020google,scott2022does}). In this paper, we have concentrated on the inverse approach but also considered a forward model.

Numerous approaches have been employed to quantitatively assess the actions of attacking players in terms of scoring, such as using expected scores derived from tracking data \cite{Lucey2014,Decroos2017,Schulte2015,Schulte2017,Bransen2018}, and action data such as dribbling and passing \cite{Decroos19,dick2021rating,yeung2023transformer}. Some researchers have valuated passes \cite{Power17,Brooks2016,dick2022can}, while others have assessed actions to receive a ball by attributing a value to the location with the highest expected score \cite{Spearman18,Link2016} and using a rule-based approach \cite{fujii2020cognition}. Defensive behaviors have also been valued using data-driven methods \cite{Robberechts2019,Toda2021}. However, such approaches would have difficulties in in considering possible (i.e., counterfactual) actions as time goes back from a goal or other events.  

From the RL perspective, numerous studies have focused on inverse approaches. To value on-ball actions, several studies have estimated Q-function or other policy functions \cite{Liu2018,Liu2020,Schulte2015,rahimian2022beyond}. However, they often consider teams as a single agent and did not valuate off-ball players in all time steps (without events). 
In terms of inverse RL, research on estimating reward functions has also been conducted \cite{Luo2020,rahimian2022inferring}. 
To estimate policy functions, researchers have sometimes performed trajectory prediction through imitation learning \cite{Le17,le2017data,Teranishi2020,fujii2020policy} and behavioral modeling \cite{Zhan19,Yeh2019,Li2021,fujii2022estimating}, aiming to mimic (rather than optimize) a policy using neural networks. 
Our approach considers the RL model overall rather than as a sub-problem using hand-crafted reward functions, and estimates multiple players’ Q-functions for simultaneously valuating on- and off-ball players even when no event occurs.

\vspace{-11pt}
\section{Conclusion}
\vspace{-6pt}
\label{sec:conclusion}
In this paper, we proposed a comprehensive evaluation method for soccer players in attacking scenes using the framework of deep RL with action supervision. 
In the experiment, we analyzed the relationship with conventional indicators, season goals, and game ratings by experts, and showed the effectiveness of the proposed method.
Our approach can assess how multiple players move continuously throughout the game, which is difficult to be discretized or labeled but vital for teamwork, scouting, and fan engagement.
A possible future direction is to improve RL models for a more valid valuation of players.
The proposed RL model and Q-value computation can be further improved by more sophisticated rewards per action (e.g., \cite{rahimian2022beyond}) and a forward RL simulation e.g., (\cite{mendes2021data}). 

\vspace{-10pt}
\section*{Acknowledgments}
\vspace{-5pt}
This work was supported by JSPS KAKENHI (Grant Numbers 20H04075 and 21H05300) and JST Presto (Grant Number JPMJPR20CA).

\bibliographystyle{splncs04}
\ifarxiv
\else
\bibliography{reference}

\begin{thebibliography}{10}
\providecommand{\url}[1]{\texttt{#1}}
\providecommand{\urlprefix}{URL }
\providecommand{\doi}[1]{https://doi.org/#1}

\bibitem{Bransen2018}
Bransen, L., Van~Haaren, J.: Measuring football players’ on-the-ball
  contributions from passes during games. In: International workshop on machine
  learning and data mining for sports analytics. pp. 3--15. Springer (2018)

\bibitem{Brooks2016}
Brooks, J., Kerr, M., Guttag, J.: Developing a data-driven player ranking in
  soccer using predictive model weights. In: Proceedings of the 22nd ACM SIGKDD
  International Conference on Knowledge Discovery and Data Mining. pp. 49--55
  (2016)

\bibitem{Decroos19}
Decroos, T., Bransen, L., Van~Haaren, J., Davis, J.: Actions speak louder than
  goals: Valuing player actions in soccer. In: KDD. pp. 1851--1861 (2019)

\bibitem{Decroos2017}
Decroos, T., Dzyuba, V., Van~Haaren, J., Davis, J.: Predicting soccer
  highlights from spatio-temporal match event streams. In: Proceedings of the
  AAAI Conference on Artificial Intelligence. vol.~31 (2017)

\bibitem{dick2022can}
Dick, U., Link, D., Brefeld, U.: Who can receive the pass? a computational
  model for quantifying availability in soccer. Data Mining and Knowledge
  Discovery  \textbf{36}(3),  987--1014 (2022)

\bibitem{dick2021rating}
Dick, U., Tavakol, M., Brefeld, U.: Rating player actions in soccer. Frontiers
  in Sports and Active Living p.~174 (2021)

\bibitem{ding2022deep}
Ding, N., Takeda, K., Fujii, K.: Deep reinforcement learning in a racket sport
  for player evaluation with technical and tactical contexts. IEEE Access
  \textbf{10},  54764--54772 (2022)

\bibitem{Fernandez18}
Fern{\'a}ndez, J., Bornn, L.: Wide open spaces: A statistical technique for
  measuring space creation in professional soccer. In: 12th MIT Sloan Sports
  Analytics Conference (2018)

\bibitem{Fernandez2019}
Fern{\'a}ndez, J., Bornn, L., Cervone, D.: Decomposing the immeasurable sport:
  A deep learning expected possession value framework for soccer. In: 13th MIT
  Sloan Sports Analytics Conference (2019)

\bibitem{fujii2021data}
Fujii, K.: Data-driven analysis for understanding team sports behaviors.
  Journal of Robotics and Mechatronics  \textbf{33}(3),  505--514 (2021)

\bibitem{fujii2020policy}
Fujii, K., Takeishi, N., Kawahara, Y., Takeda, K.: Policy learning with partial
  observation and mechanical constraints for multi-person modeling. arXiv
  preprint arXiv:2007.03155  (2020)

\bibitem{fujii2022estimating}
Fujii, K., Takeuchi, K., Kuribayashi, A., Takeishi, N., Kawahara, Y., Takeda,
  K.: Estimating counterfactual treatment outcomes over time in complex
  multi-agent scenarios. arXiv preprint arXiv:2206.01900  (2022)

\bibitem{fujii2023adaptive}
Fujii, K., Tsutsui, K., Scott, A., Nakahara, H., Takeishi, N., Kawahara, Y.:
  Adaptive action supervision in reinforcement learning from real-world
  multi-agent demonstrations. arXiv preprint arXiv:2305.13030  (2023)

\bibitem{fujii2020cognition}
Fujii, K., Yoshihara, Y., Matsumoto, Y., Tose, K., Takeuchi, H., Isobe, M.,
  Mizuta, H., Maniwa, D., Okamura, T., Murai, T., et~al.: Cognition and
  interpersonal coordination of patients with schizophrenia who have sports
  habits. PLoS One  \textbf{15}(11),  e0241863 (2020)

\bibitem{guilford1950fundamental}
Guilford, J.P.: Fundamental statistics in psychology and education. McGraw-Hill
  (1950)

\bibitem{hester2018deep}
Hester, T., Vecerik, M., Pietquin, O., Lanctot, M., Schaul, T., Piot, B.,
  Horgan, D., Quan, J., Sendonaris, A., Osband, I., et~al.: Deep q-learning
  from demonstrations. In: Proceedings of the Thirty-Second AAAI Conference on
  Artificial Intelligence and Thirtieth Innovative Applications of Artificial
  Intelligence Conference. pp. 3223--3230 (2018)

\bibitem{Score}
JLEAGUE: Jleague.jp 2019 data (2019),
  \url{https://www.jleague.jp/stats/2019/goal.html}

\bibitem{Kingma15}
Kingma, D.P., Ba, J.: Adam: A method for stochastic optimization. In:
  International Conference on Learning Representations (2015)

\bibitem{kurach2020google}
Kurach, K., Raichuk, A., Sta{\'n}czyk, P., Zaj{\k{a}}c, M., Bachem, O.,
  Espeholt, L., Riquelme, C., Vincent, D., Michalski, M., Bousquet, O., et~al.:
  Google research football: A novel reinforcement learning environment. In:
  Proceedings of the AAAI Conference on Artificial Intelligence. vol.~34, pp.
  4501--4510 (2020)

\bibitem{le2017data}
Le, H.M., Carr, P., Yue, Y., Lucey, P.: Data-driven ghosting using deep
  imitation learning. In: Proceedings of MIT Sloan Sports Analytics Conference
  (2017)

\bibitem{Le17}
Le, H.M., Yue, Y., Carr, P., Lucey, P.: Coordinated multi-agent imitation
  learning. In: Proceedings of the 34th International Conference on Machine
  Learning-Volume 70. pp. 1995--2003. JMLR. org (2017)

\bibitem{Li2021}
Li, L., Yao, J., Wenliang, L., He, T., Xiao, T., Yan, J., Wipf, D., Zhang, Z.:
  Grin: Generative relation and intention network for multi-agent trajectory
  prediction. Advances in Neural Information Processing Systems  \textbf{34},
  27107--27118 (2021)

\bibitem{Link2016}
Link, D., Lang, S., Seidenschwarz, P.: Real time quantification of dangerousity
  in football using spatiotemporal tracking data. PloS one  \textbf{11}(12),
  e0168768 (2016)

\bibitem{Liu2020}
Liu, G., Luo, Y., Schulte, O., Kharrat, T.: Deep soccer analytics: learning an
  action-value function for evaluating soccer players. Data Mining and
  Knowledge Discovery  \textbf{34}(5),  1531--1559 (2020)

\bibitem{Liu2018}
Liu, G., Schulte, O.: Deep reinforcement learning in ice hockey for
  context-aware player evaluation. In: Proceedings of the 27th International
  Joint Conference on Artificial Intelligence. pp. 3442--3448 (2018)

\bibitem{Lucey2014}
Lucey, P., Bialkowski, A., Monfort, M., Carr, P., Matthews, I.: quality vs
  quantity: Improved shot prediction in soccer using strategic features from
  spatiotemporal data. In: Proceedings of MIT Sloan Sports Analytics
  Conference. pp.~1--9 (2014)

\bibitem{Luo2020}
Luo, Y., Schulte, O., Poupart, P.: Inverse reinforcement learning for team
  sports: Valuing actions and players. In: Bessiere, C. (ed.) Proceedings of
  the Twenty-Ninth International Joint Conference on Artificial Intelligence,
  {IJCAI-20}. pp. 3356--3363. International Joint Conferences on Artificial
  Intelligence Organization (7 2020)

\bibitem{mendes2021data}
Mendes-Neves, T., Mendes-Moreira, J., Rossetti, R.J.: A data-driven simulator
  for assessing decision-making in soccer. In: Progress in Artificial
  Intelligence: 20th EPIA Conference on Artificial Intelligence, EPIA 2021,
  Virtual Event, September 7--9, 2021, Proceedings. pp. 687--698. Springer
  (2021)

\bibitem{Power17}
Power, P., Ruiz, H., Wei, X., Lucey, P.: Not all passes are created equal:
  Objectively measuring the risk and reward of passes in soccer from tracking
  data. In: KDD. pp. 1605--1613 (2017)

\bibitem{rahimian2022inferring}
Rahimian, P., Toka, L.: Inferring the strategy of offensive and defensive play
  in soccer with inverse reinforcement learning. In: Machine Learning and Data
  Mining for Sports Analytics: 8th International Workshop, MLSA 2021, Virtual
  Event, September 13, 2021, Revised Selected Papers. pp. 26--38. Springer
  (2022)

\bibitem{rahimian2022beyond}
Rahimian, P., Van~Haaren, J., Abzhanova, T., Toka, L.: Beyond action valuation:
  A deep reinforcement learning framework for optimizing player decisions in
  soccer. In: 16th Annual MIT Sloan Sports Analytics Conference. Boston, MA,
  USA: MIT. p.~25 (2022)

\bibitem{Robberechts2019}
Robberechts, P.: Valuing the art of pressing. In: Proceedings of the StatsBomb
  Innovation In Football Conference. pp. 1--11. StatsBomb (2019)

\bibitem{Schulte2015}
Routley, K., Schulte, O.: A markov game model for valuing player actions in ice
  hockey. In: Proceedings of the Thirty-First Conference on Uncertainty in
  Artificial Intelligence. p. 782–791. UAI'15, AUAI Press, Arlington,
  Virginia, USA (2015)

\bibitem{Schulte2017}
Schulte, O., Khademi, M., Gholami, S., Zhao, Z., Javan, M., Desaulniers, P.: A
  markov game model for valuing actions, locations, and team performance in ice
  hockey. Data Mining and Knowledge Discovery  \textbf{31}(6),  1735--1757
  (2017)

\bibitem{scott2022does}
Scott, A., Fujii, K., Onishi, M.: How does {AI} play football? {A}n analysis of
  {RL} and real-world football strategies. In: 14th International Conference on
  Agents and Artificial Intelligence (ICAART' 22). vol.~1, pp. 42--52 (2022)

\bibitem{Rating}
Soccer-digest: Soccer digest web j1 rating (2019),
  \url{https://www.soccerdigestweb.com}

\bibitem{Spearman18}
Spearman, W.: Beyond expected goals. In: Proceedings of the 12th MIT sloan
  sports analytics conference. pp. 1--17 (2018)

\bibitem{Teranishi2020}
Teranishi, M., Fujii, K., Takeda, K.: Trajectory prediction with imitation
  learning reflecting defensive evaluation in team sports. In: 2020 IEEE 9th
  Global Conference on Consumer Electronics (GCCE). pp. 124--125. IEEE (2020)

\bibitem{teranishi2022evaluation}
Teranishi, M., Tsutsui, K., Takeda, K., Fujii, K.: Evaluation of creating
  scoring opportunities for teammates in soccer via trajectory prediction. In:
  International Workshop on Machine Learning and Data Mining for Sports
  Analytics. Springer (2022)

\bibitem{Toda2021}
Toda, K., Teranishi, M., Kushiro, K., Fujii, K.: Evaluation of soccer team
  defense based on prediction models of ball recovery and being attacked. PLoS
  One  \textbf{17}(1),  e0263051 (2022)

\bibitem{umemoto2022location}
Umemoto, R., Tsutsui, K., Fujii, K.: Location analysis of players in uefa euro
  2020 and 2022 using generalized valuation of defense by estimating
  probabilities. arXiv preprint arXiv:2212.00021  (2022)

\bibitem{Yeh2019}
Yeh, R.A., Schwing, A.G., Huang, J., Murphy, K.: Diverse generation for
  multi-agent sports games. In: The IEEE Conference on Computer Vision and
  Pattern Recognition (CVPR) (June 2019)

\bibitem{yeung2023transformer}
Yeung, C.C.K., Sit, T., Fujii, K.: Transformer-based neural marked spatio
  temporal point process model for football match events analysis. arXiv
  preprint arXiv:2302.09276  (2022)

\bibitem{Zhan19}
Zhan, E., Zheng, S., Yue, Y., Sha, L., Lucey, P.: Generating multi-agent
  trajectories using programmatic weak supervision. In: International
  Conference on Learning Representations (2019)

\end{thebibliography}
\fi

\end{document}